\documentclass[letterpaper, 10 pt, conference]{ieeeconf}

\IEEEoverridecommandlockouts  
\def\endthebibliography{%
  \def\@noitemerr{\@latex@warning{Empty `thebibliography' environment}}%
  \endlist
}


\usepackage{color}

\title{\LARGE \bf
    Human-Guided Planner for Non-Prehensile Manipulation
}

\author{Rafael Papallas and Mehmet R. Dogar
    \thanks{This research has received funding from the European Union's Horizon 2020 research
        and innovation programme under the Marie Sklodowska-Curie grants agreement
        No. 746143, and from the UK Engineering and Physical Sciences Research Council
        under grant EP/N509681/1, EP/P019560/1 and EP/R031193/1.}
\thanks{Authors are with the School of Computing, University of Leeds, United Kingdom
        {\tt\small \{r.papallas, m.r.dogar\}@leeds.ac.uk}}%
}

\usepackage{tabularx,booktabs}
\newcolumntype{C}{>{\centering\arraybackslash}X} 
\usepackage{hhline}
\usepackage{multirow}

\usepackage{caption}
\usepackage{subcaption}
\usepackage{graphicx,xcolor}

\usepackage{algorithm}
\usepackage[noend]{algpseudocode}
\algnewcommand{\IIf}[1]{\State\algorithmicif\ #1\ \algorithmicthen}
\algnewcommand{\EndIIf}{\unskip\ \algorithmicend\ \algorithmicif}
\algdef{SE}[DOWHILE]{Do}{doWhile}{\algorithmicdo}[1]{\algorithmicwhile\ #1}%

\usepackage{mathtools}
\usepackage{bm}
\usepackage{esvect}
\usepackage{amssymb}
\usepackage{xspace}

\usepackage{scalerel}

\usepackage[english]{babel}

\usepackage[utf8]{inputenc}

\usepackage{csquotes}

\usepackage{graphicx,import}
\graphicspath{{./}}

\usepackage{booktabs}

\usepackage[per-mode=symbol]{siunitx}

\usepackage{placeins}

\usepackage{hyperref}

\usepackage[capitalize]{cleveref}
\Crefname{algorithm}{Alg.}{Algs.}
\Crefname{section}{Sec.}{Secs}
\Crefname{line}{line}{lines}

\usepackage{microtype}
\usepackage[noadjust]{cite}
\newcommand{\acronymourapproach}{GRTC\xspace}

\newcommand{\acronymbaseline}{RTC\xspace}
\newcommand{\lreaching}{reaching through clutter\xspace}

\newcommand{\ourplanneracronym}{\acronymourapproach-HITL\xspace}
\newcommand{\heuristicplannername}{\acronymourapproach-Heuristic\xspace}
\newcommand{\ourplannerinfullacronym}{Guided-RTC with Human-In-The-Loop (\ourplanneracronym)\xspace}

\newcommand{\initialState}{q_{0}}
\newcommand{\goalStates}{Q_{goals}}
\newcommand{\goalState}{q_{n}}
\newcommand{\stateSpace}{Q}
\newcommand{\controlSpace}{U}

\newcommand{\systemDynamics}{f}
\newcommand{\systemDynamicsFunction}{{\systemDynamics: \stateSpace \times \controlSpace \to \stateSpace}}

\newcommand{\object}[1]{o_{#1}}
\newcommand{\guideState}[1]{{(\object{#1}, x_{#1}, y_{#1})}}

\begin{document}
    \newcommand{\testing}{Feedback and Revaluation}

    \maketitle
    \thispagestyle{empty}
    \pagestyle{empty}

    \begin{abstract}
We present a human-guided planner for non-prehensile manipulation in clutter.
Most recent approaches to manipulation in clutter employs randomized planning, however, the problem
remains a challenging one where the planning times are still in the order of tens of seconds or minutes, and the
success rates are low for difficult instances of the problem. We build on these
control-based randomized planning approaches, but we investigate using them in conjunction
with human-operator input. We show that with a minimal amount of human input, the low-level planner can solve the
problem faster and with higher success rates.
\end{abstract}

    \section{Introduction}
    \label{sec:introduction}

\begin{figure}[!t]
        \vspace*{2mm}
        \captionsetup[subfigure]{aboveskip=1.0pt,belowskip=1.0pt}
        \centering
        \begin{subfigure}{.33\linewidth}
            \centering
            \tiny
            \def\svgwidth{0.85\columnwidth}
\begingroup%
  \makeatletter%
  \providecommand\color[2][]{%
    \errmessage{(Inkscape) Color is used for the text in Inkscape, but the package 'color.sty' is not loaded}%
    \renewcommand\color[2][]{}%
  }%
  \providecommand\transparent[1]{%
    \errmessage{(Inkscape) Transparency is used (non-zero) for the text in Inkscape, but the package 'transparent.sty' is not loaded}%
    \renewcommand\transparent[1]{}%
  }%
  \providecommand\rotatebox[2]{#2}%
  \newcommand*\fsize{\dimexpr\f@size pt\relax}%
  \newcommand*\lineheight[1]{\fontsize{\fsize}{#1\fsize}\selectfont}%
  \ifx\svgwidth\undefined%
    \setlength{\unitlength}{950.00006104bp}%
    \ifx\svgscale\undefined%
      \relax%
    \else%
      \setlength{\unitlength}{\unitlength * \real{\svgscale}}%
    \fi%
  \else%
    \setlength{\unitlength}{\svgwidth}%
  \fi%
  \global\let\svgwidth\undefined%
  \global\let\svgscale\undefined%
  \makeatother%
  \begin{picture}(1,0.94736836)%
    \lineheight{1}%
    \setlength\tabcolsep{0pt}%
    \put(0,0){\includegraphics[width=\unitlength,page=1]{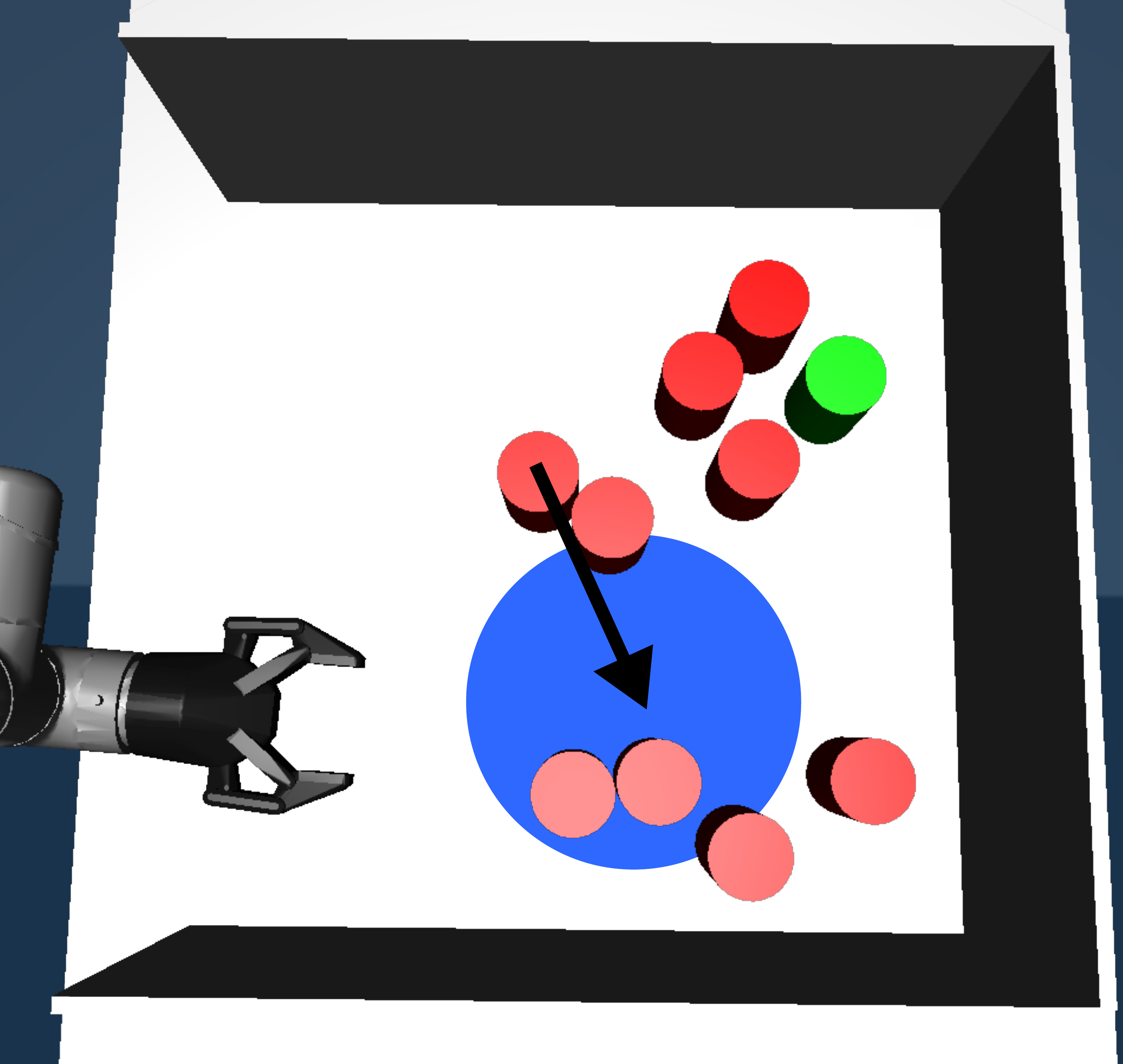}}%
    \put(0.45812889,0.57002671){\color[rgb]{0,0,0}\makebox(0,0)[lt]{\lineheight{1.25}\smash{\begin{tabular}[t]{l}$o_2$\end{tabular}}}}%
    \put(0.7494225,0.653492){\color[rgb]{0,0,0}\makebox(0,0)[lt]{\lineheight{1.25}\smash{\begin{tabular}[t]{l}$o_g$\end{tabular}}}}%
  \end{picture}%
\endgroup%

            \caption{}
            \label{fig:guidance_explained_1}
        \end{subfigure}%
        \begin{subfigure}{.33\linewidth}
            \centering
            \tiny
            \def\svgwidth{0.85\columnwidth}
            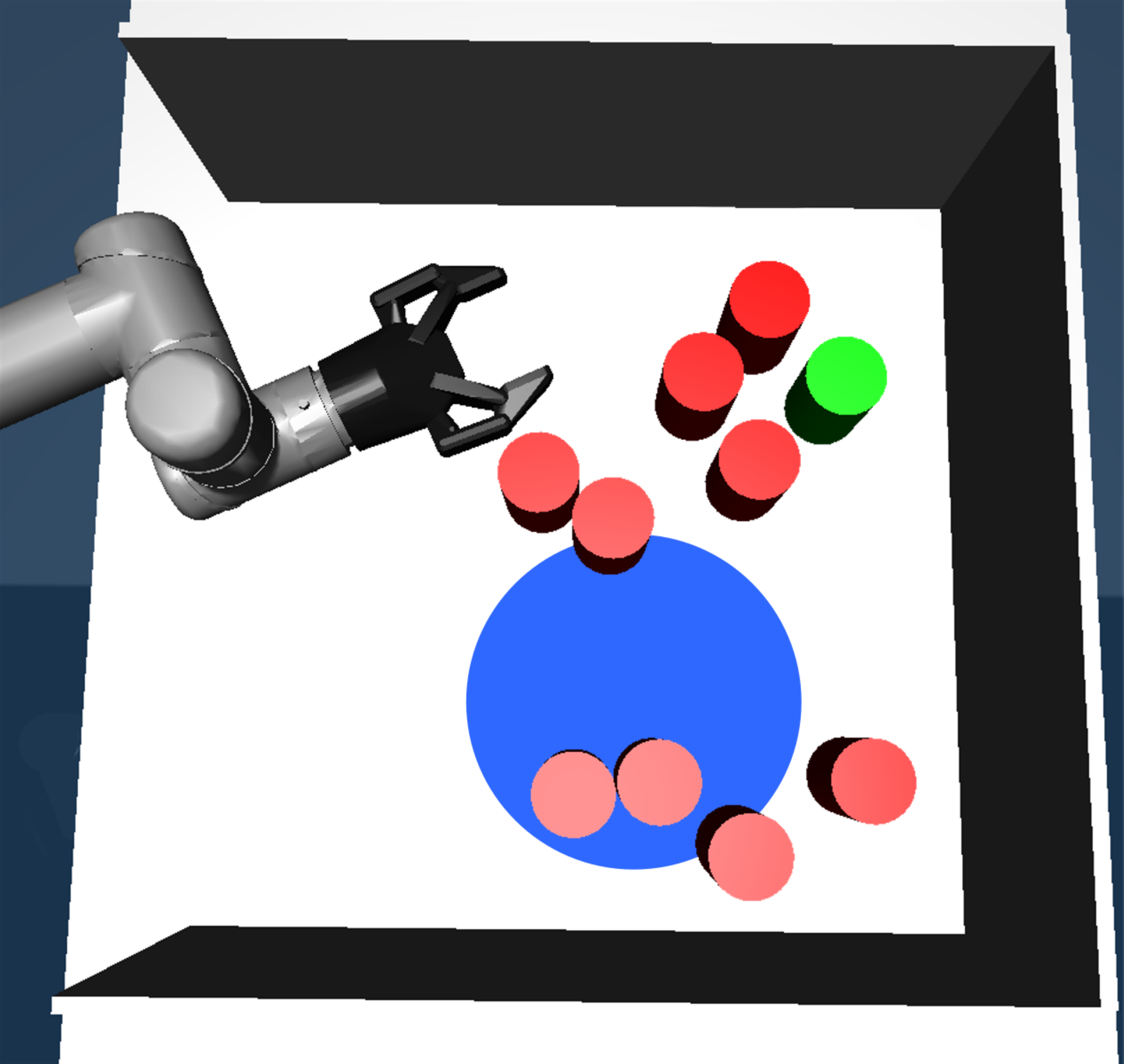
            \caption{}
            \label{fig:guidance_explained_2}
        \end{subfigure}%
        \begin{subfigure}{.33\linewidth}
            \centering
            \tiny
            \def\svgwidth{0.85\columnwidth}
            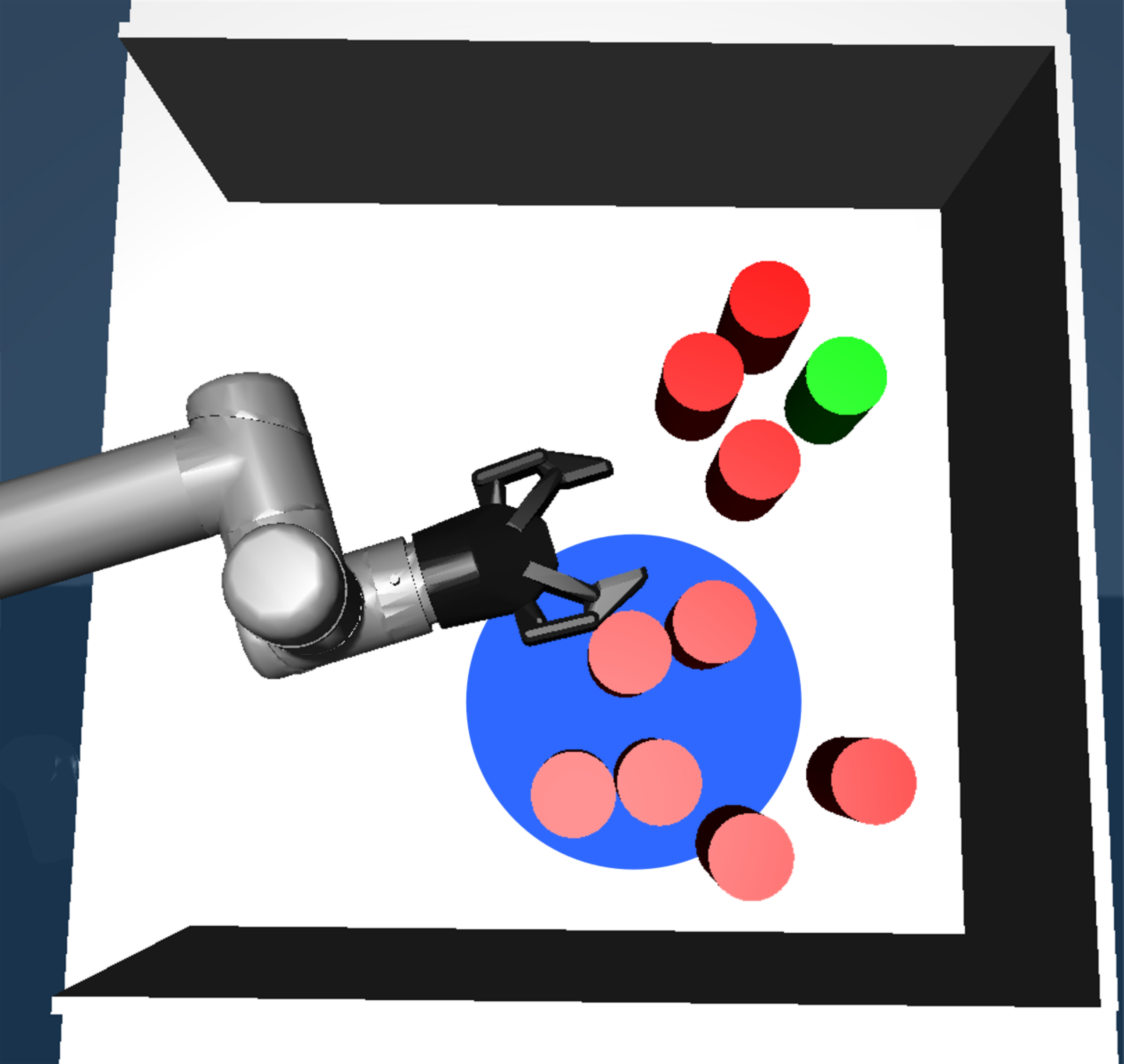
            \caption{}
            \label{fig:guidance_explained_3}
        \end{subfigure}

        \begin{subfigure}{.33\linewidth}
            \centering
            \tiny
            \def\svgwidth{0.85\columnwidth}
            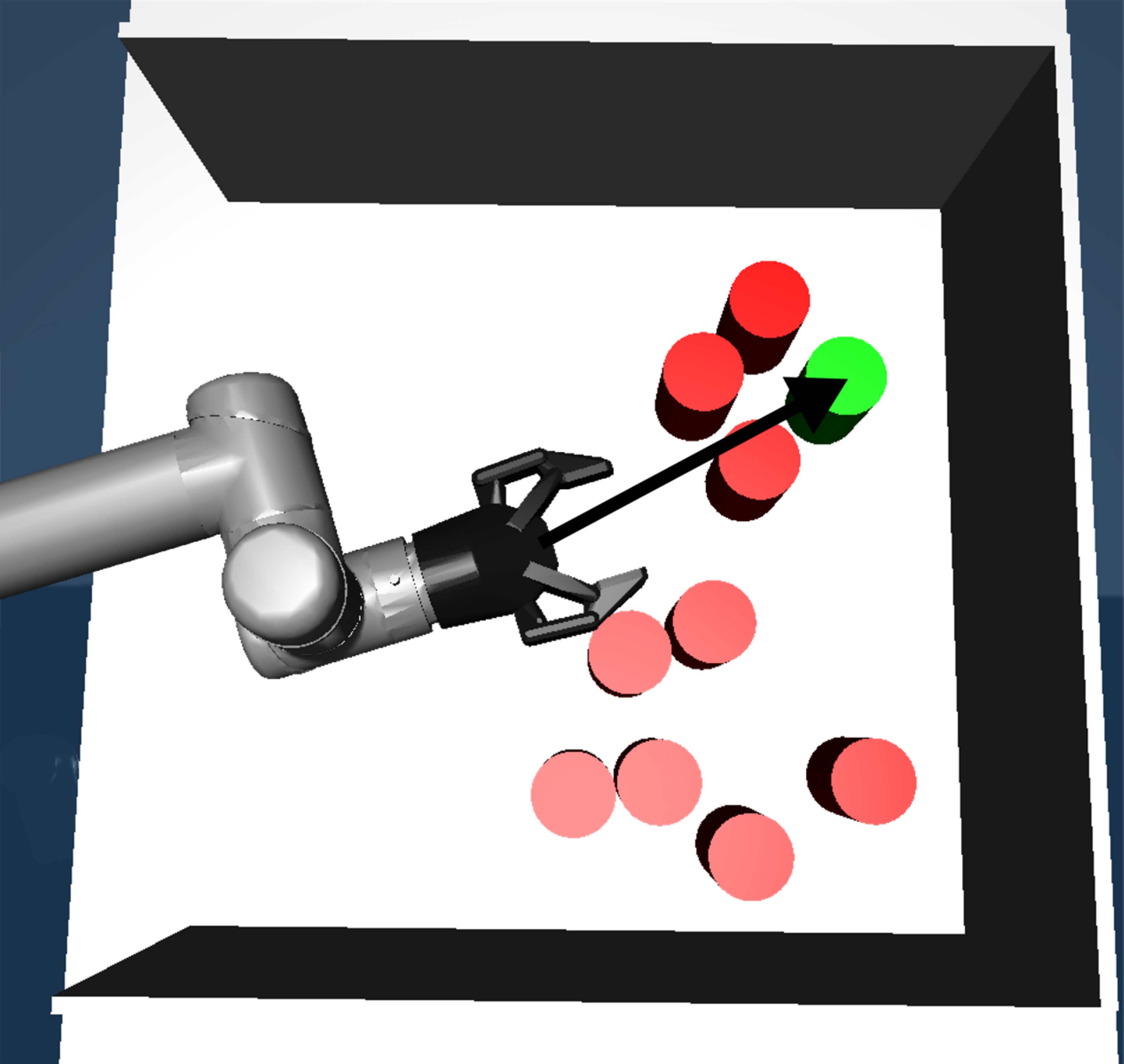
            \caption{}
            \label{fig:guidance_explained_4}
        \end{subfigure}%
        \begin{subfigure}{.33\linewidth}
            \centering
            \tiny
            \def\svgwidth{0.85\columnwidth}
            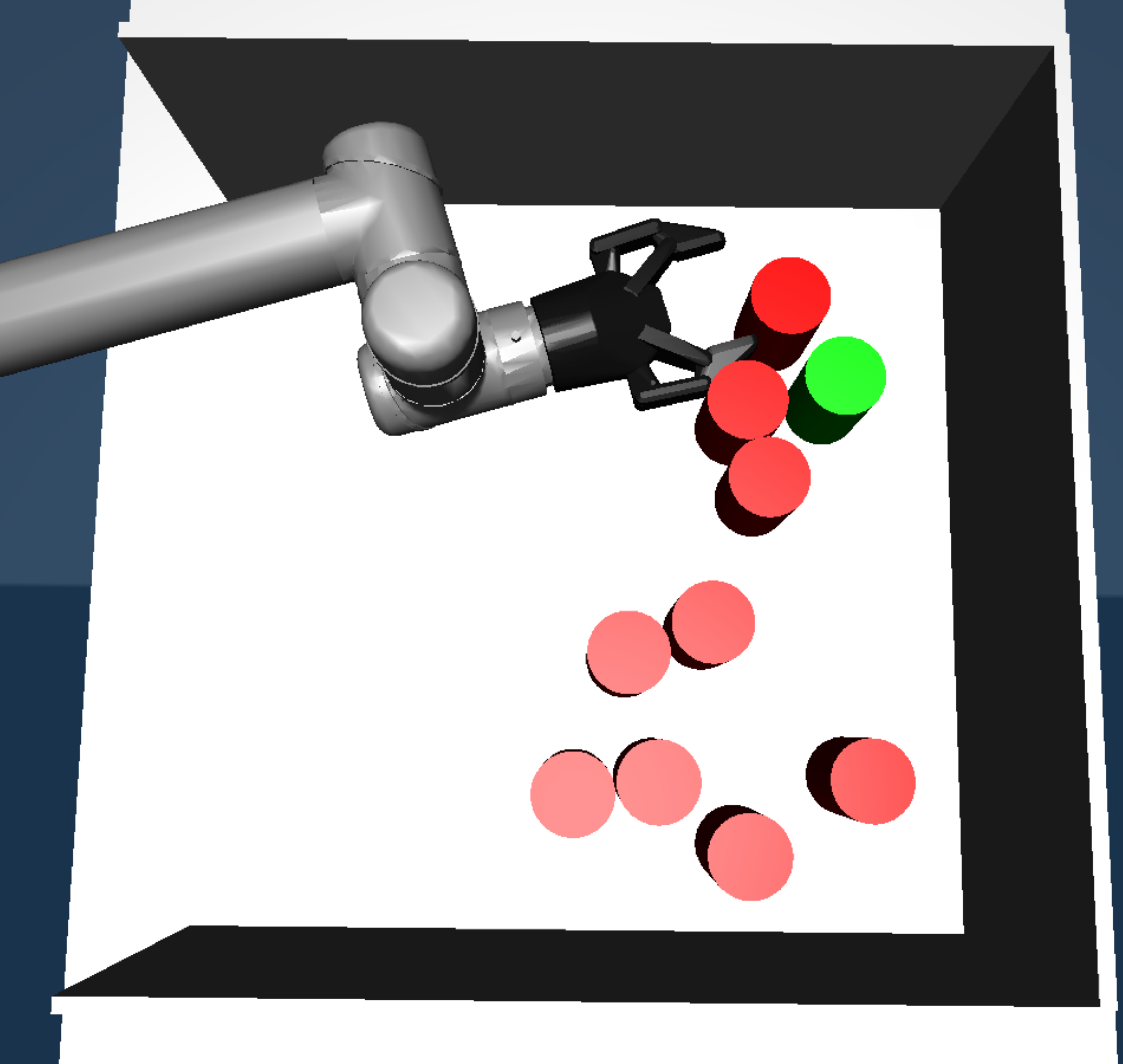
            \caption{}
            \label{fig:guidance_explained_5}
        \end{subfigure}%
        \begin{subfigure}{.325\linewidth}
            \centering
            \tiny
            \def\svgwidth{0.85\columnwidth}
            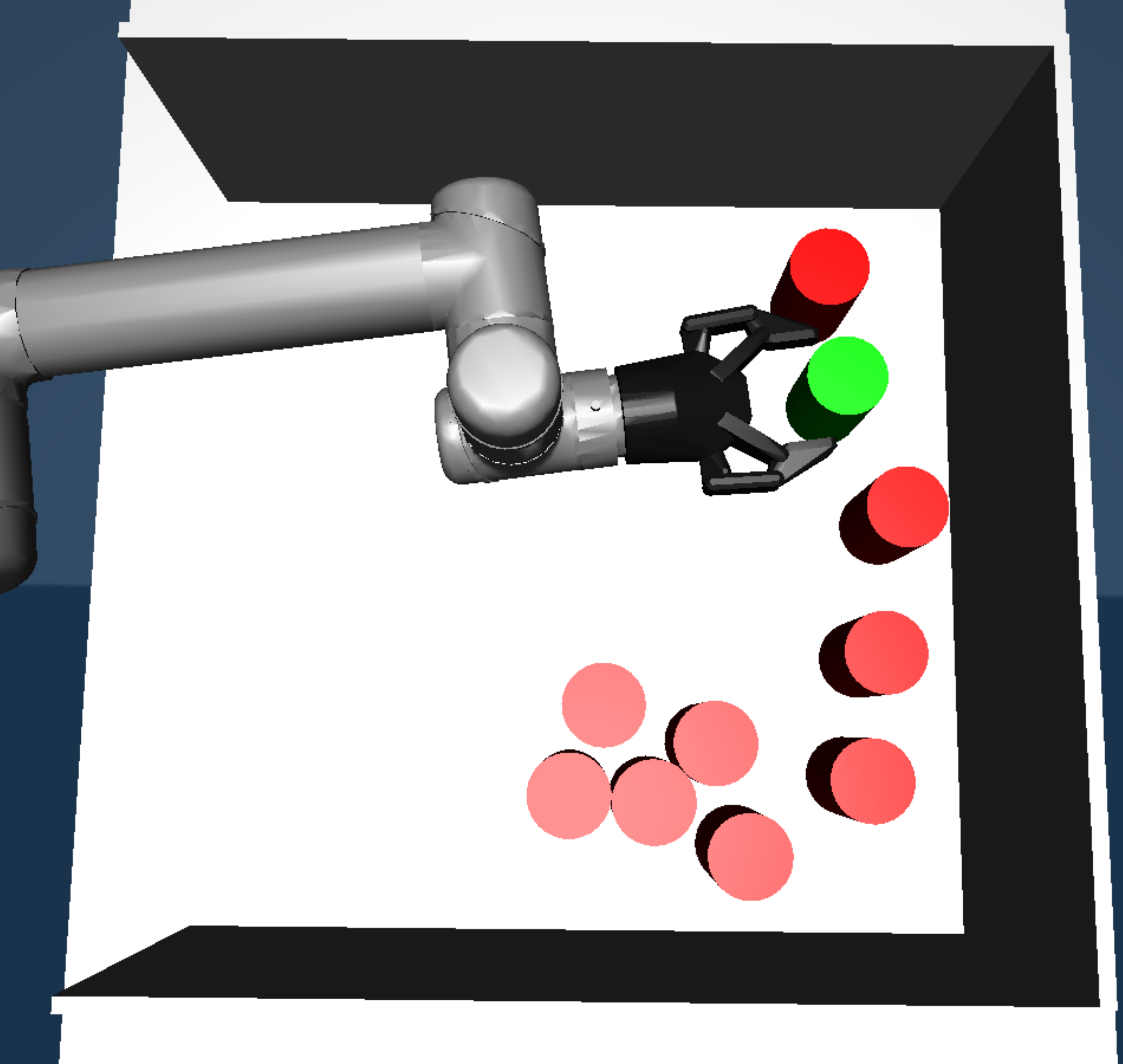
            \caption{}
            \label{fig:guidance_explained_6}
        \end{subfigure}%
        \caption{A human-operator guiding a robot to reach for the green goal
        object, $o_g$. Arrows indicate human interaction with the robot.
        In (a) the operator indicates $o_2$ to be pushed to
        the blue target region. From (a) to (c) the robot plans to
        perform this push. In (d) the operator indicates to the robot
        to reach for the goal object. From (d) to (f) the robot plans to reach
        the goal object.}
        \label{fig:guidance_explained}
    \end{figure}

\begin{figure}[!b]
        \captionsetup[subfigure]{aboveskip=1.0pt,belowskip=1.0pt}
        \centering
        \begin{subfigure}{.25\linewidth}
            \centering
            \def\svgwidth{0.95\columnwidth}
            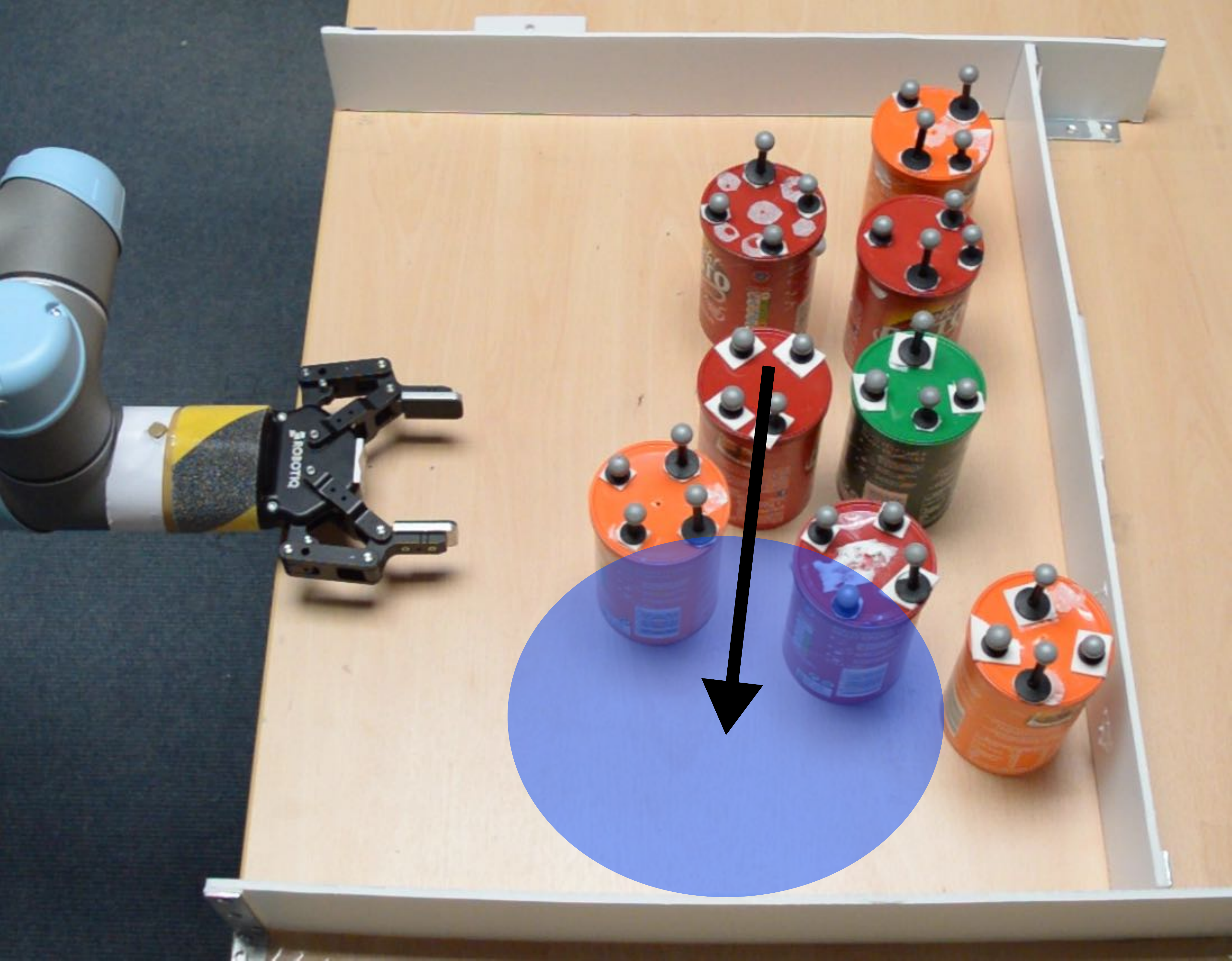
            \caption{}
            \label{fig:real_experiment_explained_1_1}
        \end{subfigure}%
        \begin{subfigure}{.25\linewidth}
            \centering
            \def\svgwidth{0.95\columnwidth}
            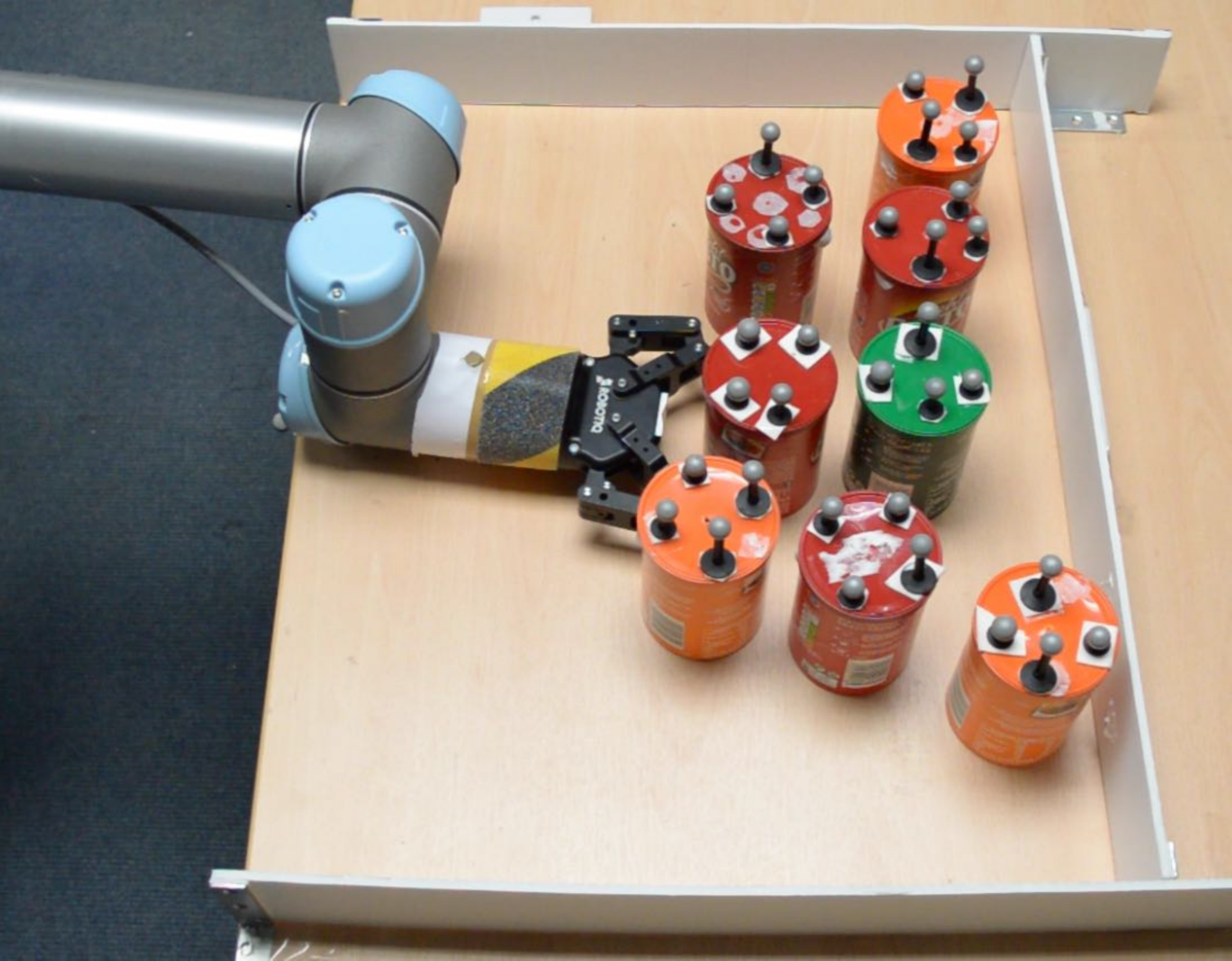
            \caption{}
            \label{fig:real_experiment_explained_1_2}
        \end{subfigure}%
        \begin{subfigure}{.25\linewidth}
            \centering
            \def\svgwidth{0.95\columnwidth}
            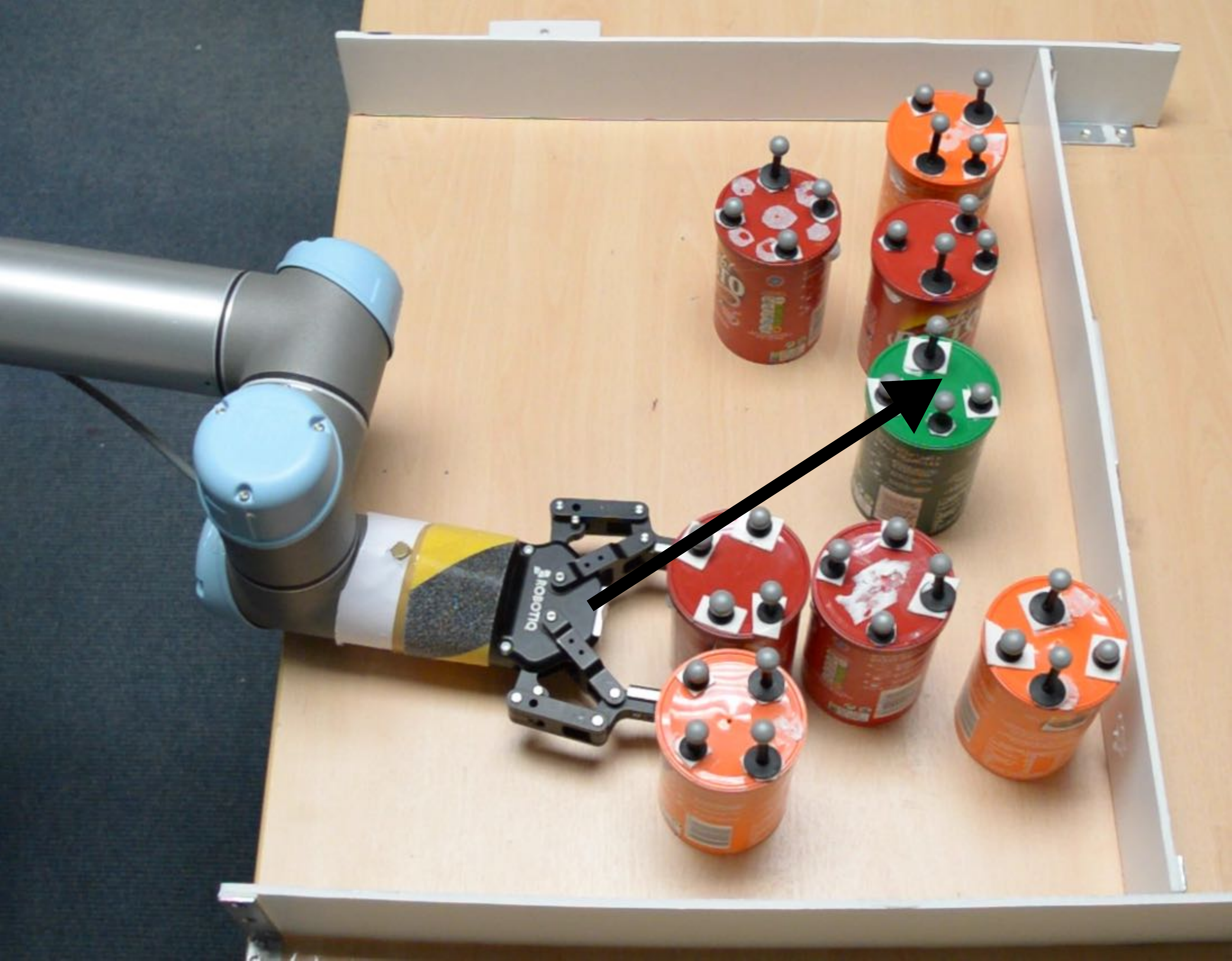
            \caption{}
            \label{fig:real_experiment_explained_1_3}
        \end{subfigure}%
        \begin{subfigure}{.25\linewidth}
            \centering
            \def\svgwidth{0.95\columnwidth}
            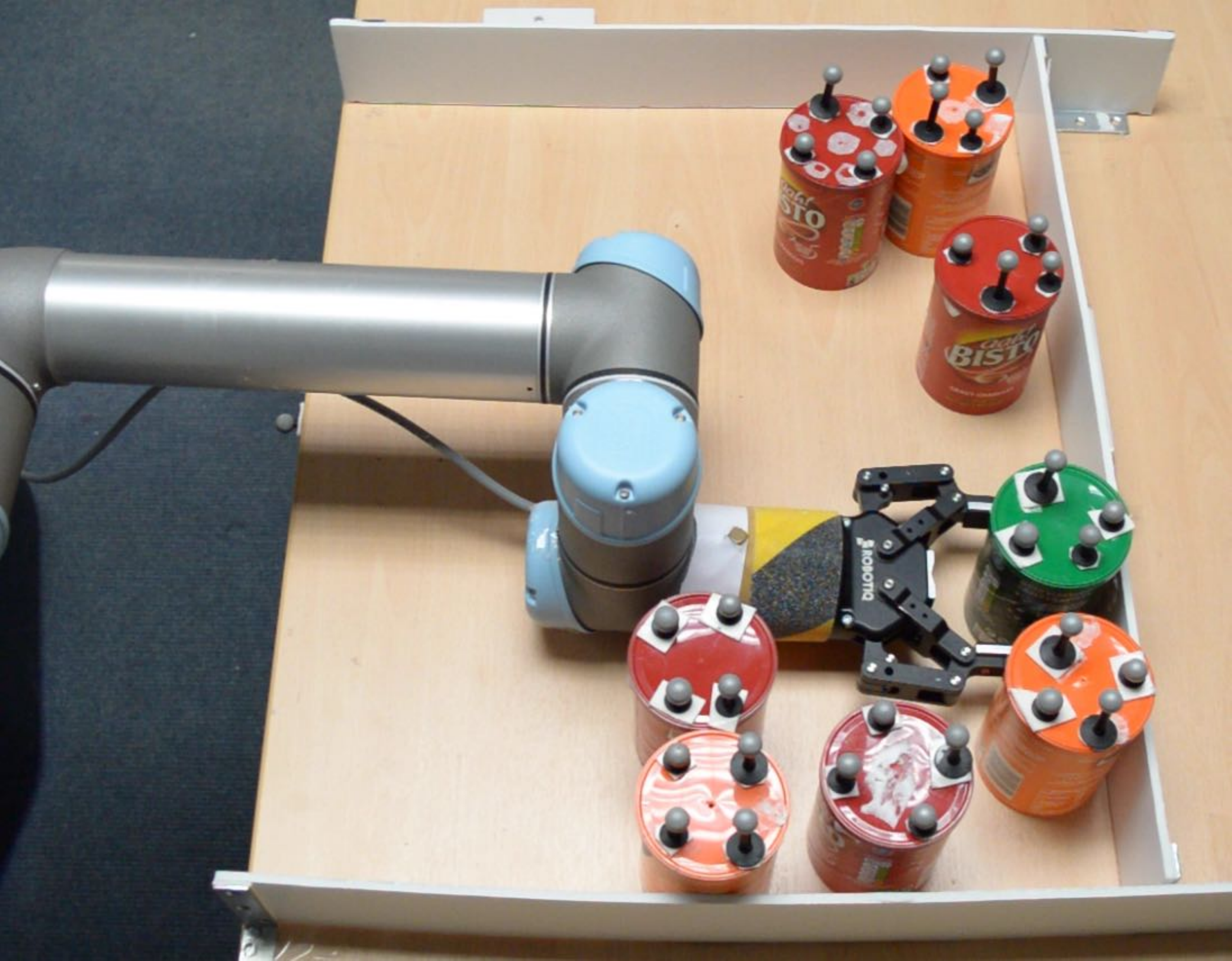
            \caption{}
            \label{fig:real_experiment_explained_1_4}
        \end{subfigure}
        \caption{Human-operator guiding a robot in the real-world.}
        \label{fig:real_experiment_explained_1}
    \end{figure}

    We present a human-guided planner for non-prehensile
    manipulation in clutter. We show example problems in
    \cref{fig:guidance_explained,fig:real_experiment_explained_1}. The target
    of the robot is to reach and grasp the green object. To do this,
    however, the robot first has to push other objects out of
    the way (\cref{fig:guidance_explained_2} to
    \cref{fig:guidance_explained_5}). This requires the robot to
    plan which objects to contact, where and how to push those objects so
    that it can reach the goal object.

    These \textit{\lreaching} problems are difficult to solve due
    to several reasons: First, the number of objects make the state space of
    high-dimensionality. Second, this is an underactuated problem, since the
    objects cannot be controlled by the robot directly. Third,
    predicting the evolution of the system state requires running computationally
    expensive physics simulators, to predict how objects would move as a result of the robot pushing. Effective algorithms have been developed
    \cite{dogar2012physics,havur2014geometric,kitaev2015physics,
    haustein2015kinodynamic,moll2018randomized,bejjani2018,bejjani2019learning,
    king2016rearrangement,agboh2018real,huang2019large,kim2019retrieving},
    however the problem remains a challenging one, where the planning times
    are still in the order of tens of seconds or minutes, and the success rates are
    low for difficult problems.

    Further study of the \lreaching problem is important to develop
    approaches to solve the problem more successfully and faster. It
    is a problem that has a potential for major near-term impact in warehouse
    robotics and personal home robots. The algorithms that we currently have, however, are not able to solve
    \lreaching problems in the real world in a fast and consistent way.
    Here, we ask the question of whether human-operators can be used to provide
    a minimal amount of input that results in a significantly higher success
    rate and faster planning times.

    Most recent approaches to the \lreaching problem employs the power of randomized
    \textit{kinodynamic planning}. Haustein et al.
    \cite{haustein2015kinodynamic} use a kinodynamic RRT
    \cite{lavalle1998rrt,lavalle2001randomized} planner to sample and generate a
    sequence of robot pushes on objects to reach a goal state. Muhayyuddin et al.
    \cite{moll2018randomized} use the KPIECE algorithm \cite{csucan2009kpiece} to
    solve this problem. These planners report some of the best performance (in terms of
    planning times and success rates) in this domain so far.

    We build on these kinodynamic planning approaches,
    but we investigate using them in conjunction with human-operator input. In our
    framework, the human operator supplies a \textit{high-level plan} to make the
    underlying planners solve the problem faster and with higher success rates.

    \section{Problem Formulation}%
    \label{sec:formulation}
    Our environment is comprised of a robot $r$, a set of movable obstacles $O$, and
    other static obstacles. The robot is allowed to interact with the movable obstacles, but not with the static ones.
    We also have ${o_g \in O}$ which is the \textit{goal} object to reach.

    We are interested in problems where the robot needs to reach for an 
    object in a cluttered shelf that is constrained from the top, and
    therefore we constrain the robot motion to the plane
    and its configuration space, $Q^r$, to $SE(2)$. The configuration of a movable 
    object ${i \in \{1, \dots, |O|\}}$, $q^i$, is its pose on the plane
    ($x, y, \theta$). We denote its configuration space as $Q^{i}$.
    The configuration space of the complete system is the Cartesian product
    ${Q = Q^r \times Q^{g} \times Q^{1} \times \dots \times Q^{|O|-1}}$.

    Let ${\initialState \in \stateSpace}$ be the initial
    configuration of the system, and ${\goalStates \subset \stateSpace}$ a set of possible
    goal configurations. A goal configuration, ${\goalState \in Q_{goals}}$, is defined as a configuration where $o_g$ is within
    the robot's end-effector (see \cref{fig:guidance_explained_6}).

    Let $\controlSpace$ be the control space comprised of the robot
    velocities. Let the system dynamics be defined as $\systemDynamicsFunction$ that propagates
    the system from ${q_t \in \stateSpace}$ with a control
    ${u_t \in \controlSpace}$. 
    
    We define the \textit{Reaching Through Clutter} (\acronymbaseline) problem as the tuple
    ${(\stateSpace, \controlSpace, \initialState, \goalStates, \systemDynamics)}$. The solution 
    to the problem is a sequence of
    controls from $\controlSpace$ that move
    the robot from $\initialState$ to a ${\goalState \in Q_{goals}}$.

    \section{Sampling-based Kinodynamic Planners}%
    \label{sec:rtc_planners}

    Two well known sampling-based kinodynamic planners are Rapidly-exploring Random
    Trees (RRT) \cite{lavalle1998rrt, lavalle2001randomized} and
    Kinodynamic Motion Planning by Interior-Exterior Cell Exploration (KPIECE)
    \cite{csucan2009kpiece}.
    We use kinodynamic RRT and KPIECE in our work in two different ways: (1) as
    baseline RTC planners to compare against, and (2) as the low-level planners
    for the Guided-RTC Framework that accepts high-level actions (explained in \cref{sec:guidance_planner}).

    In this work, when we plan with a kinodynamic planner (either RRT or KPIECE) we will
    use the notation \textit{kinodynamicPlanning($q_{start}$, goal)} with a
    start configuration of the system, $q_{start}$, and some \textit{goal} input.

\section{Guided-RTC Framework}%
    \label{sec:guidance_planner}

    In this section we describe a \textit{guided} system to solve RTC problems.
    A Guided-RTC system accepts high-level actions.
    A high-level action can suggest to push a particular
    obstacle object into a certain region, or it may suggest to reach for the goal
    object. We formally define a high-level action with the triple $\guideState{i}$, where
    ${o_i \in O}$ is an object, and ${(x_i, y_i)}$ is the centroid of
    a \textit{target region} that $o_i$ needs to be pushed into. The target
    region has a constant diameter $d$.

    In this work, we investigate how a Guided-RTC system
    with a human-in-the-loop performs when compared with (a) solving the original RTC problem
    directly using kinodynamic approaches (\cref{sec:rtc_planners}), and (b)
    using Guided-RTC systems with automated ways of generating the high-level
    actions.

    \subsection{A Generic approach for Guided-RTC Planning}%
        \label{sub:grtc_planner}

        We present a generic algorithm for Guided-RTC in \cref{alg:grtc}.
        The next high-level action is decided based on the current
        configuration (\cref{algline:get_next_action}). If the object in the
        high-level action is not the goal object (\cref{algline:ifnotgoal}),
        then it is pushed to the target region between
        \cref{algline:approaching_states,algline:first_execution},
        and a new high-level action is requested.
        If it is the goal object, the robot tries to reach it between
        \cref{algline:reach_goal,algline:second_execution} and the
        system terminates.

\begin{figure}[!t]
            \vspace*{2mm}
            \tiny
            \centering
            \def\svgwidth{0.9\columnwidth}
            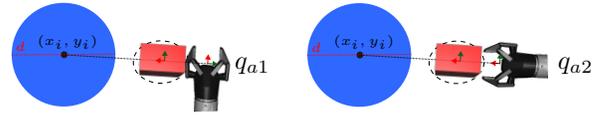
            \caption{Approaching states: The blue circle is the target region, the red rectangle the object to
            manipulate. We compute two approaching states, $q_{a1}$ and $q_{a2}$.}
            \label{fig:approaching_explained}
        \end{figure}

        We plan to push an object to its target region in two steps. On
        \cref{algline:first_kinodynamic_planning} we plan to an intermediate
        \textit{approaching state} near the object, and then on
        \cref{algline:second_kinodynamic_planning}, we plan from this
        approaching state to push the object to its target region.
        Specifically, given an object to push, $o_i$, we compute two
        approaching states $q_{a1}$ and $q_{a2}$
        (\cref{algline:approaching_states}).
        \cref{fig:approaching_explained} shows how these approaching
        states are computed, based on the object's current
        position, the centroid $(x_i, y_i)$ and the minimum enclosing circle
        of the object. The approaching state $q_{a1}$ encourages side-ways
        pushing, where $q_{a2}$ encourages forward pushing.
        We also experimented with planning
        without first approaching the object but we found that
        approaching the object from a good pose yields to faster pushing
        solutions.
        Using both approaching states as the goal we plan to move to one of them
        (multi-goal planning) on \cref{algline:first_kinodynamic_planning}.
        Then, from the approaching state reached (either $q_{a1}$ or $q_{a2}$) we
        push $o_i$ to its target region (\cref{algline:second_kinodynamic_planning}).
        If any of the two planning calls on \cref{algline:first_kinodynamic_planning,algline:second_kinodynamic_planning} fails, then the algorithm proceeds to the
        next high-level action (\cref{algline:get_next_action}). Otherwise,
        we execute the solutions sequentially on
        \cref{algline:first_execution}, which changes the current system configuration $q_{current}$.

        \begin{algorithm}[!b]
			\caption{Guided-RTC}\label{alg:grtc}
			\algrenewcommand\algorithmicindent{0.8em}%
			\begin{algorithmic}[1]
				\Procedure{GRTC}{$\stateSpace, \controlSpace, \initialState, \goalStates$}
                \State $q_{current} \gets q_0$ \label{algline:qcurrent}
					\Do
                        \State $o_i, x_i, y_i \gets$ \textproc{NextHighLevelAction($q_{current}$)} \label{algline:get_next_action}
                        \If{$o_i \neq o_g$} \label{algline:ifnotgoal}
                            \State $q_{a1}, q_{a2} \gets$ compute approaching states to $o_i$ \label{algline:approaching_states}
                            \State kinodynamicPlanning($q_{current}, \{q_{a1}, q_{a2}\}$) \label{algline:first_kinodynamic_planning}

                            \State \textbf{if} planning fails \textbf{then} \textbf{continue}

                            \State kinodynamicPlanning($q_{a1}$ or $q_{a2}, (o_i, {x_i, y_i})$) \label{algline:second_kinodynamic_planning}
                            \State \textbf{if} planning fails \textbf{then} \textbf{continue}
                            \State $q_{current} \gets$ execute solutions from \cref{algline:first_kinodynamic_planning,algline:second_kinodynamic_planning} \label{algline:first_execution}
						\EndIf
					\doWhile{$o_i \neq o_g$}
                    \State kinodynamicPlanning($q_{current}, \goalStates$) \label{algline:reach_goal}
					\If{planning succeeds}
                        \State $q_{current} \gets$ execute solution from \cref{algline:reach_goal} \label{algline:second_execution}
					\EndIf
				\EndProcedure
			\end{algorithmic}
		\end{algorithm}

        \cref{alg:grtc} runs up to an overall time limit, $T_{overall}$, or
        until a goal is reached. The pushing planning calls on
        \cref{algline:first_kinodynamic_planning,algline:second_kinodynamic_planning}
        have their own shorter time limit, $T_{pushing}$, and they should find a
        valid solution within this limit.
        The planning call on \cref{algline:reach_goal} is allowed to
        run until the overall time limit is over.

    \subsection{\ourplannerinfullacronym}%
        \label{sub:grtc_hitl_planner}

        \begin{algorithm}[!t]
			\caption{\ourplanneracronym}\label{alg:grtc_hitl}
			\algrenewcommand\algorithmicindent{0.8em}%
			\begin{algorithmic}[1]
                \Function{NextHighLevelAction}{$q_{current}$}
					\State $o_i \gets$ get object selection from human operator
					\If{$o_i \neq o_g$}
						\State $x_i, y_i \gets$ get region centroid from human operator
						\State \textbf{return} $o_i, x_i, y_i$
					\EndIf
					\State \textbf{return} $o_g$
				\EndFunction
			\end{algorithmic}
		\end{algorithm}

        \ourplannerinfullacronym is an instantiation of the \acronymourapproach
        Framework. A human-operator, through a graphical user interface, provides the
		high-level actions. In \cref{alg:grtc_hitl} we present
        \ourplanneracronym \textproc{NextHighLevelAction} function
        (referenced in \cref{alg:grtc}, \cref{algline:get_next_action}).

		The human provides high-level actions until she selects the goal
		object, $o_g$. The \acronymourapproach framework (\cref{alg:grtc})
		plans and executes them. The state of the system changes after each
		high-level action and the human operator is presented with the resulting state
        each time ($q_{current}$). Note here that \textit{the operator can decide not to
        provide any guidance}.

        We developed a simple user interface to communicate with the human-operator.
        The operator, using a mouse pointer, provides the input by first clicking
		on the desired object and then a point on the plane (\cref{fig:guidance_explained_1})
        that becomes the centroid of the target region.

        The approach we propose here uses a human-operator to
        decide on the high-level plan. One question is whether one
        can use automatic approaches, and how they would perform
        compared to the human suggested actions. To make such a
        comparison, we implemented an automated approach (\heuristicplannername, \cref{sub:grtc_heuristic_planner}).

    \subsection{Guided-RTC with Straight Line Heuristic (\heuristicplannername)}%
        \label{sub:grtc_heuristic_planner}

        \begin{figure}[!b]
            \captionsetup[subfigure]{aboveskip=1.0pt,belowskip=1.0pt}
            \centering
            \begin{subfigure}{.3\linewidth}
                \centering
                \tiny
                \def\svgwidth{0.9\columnwidth}
                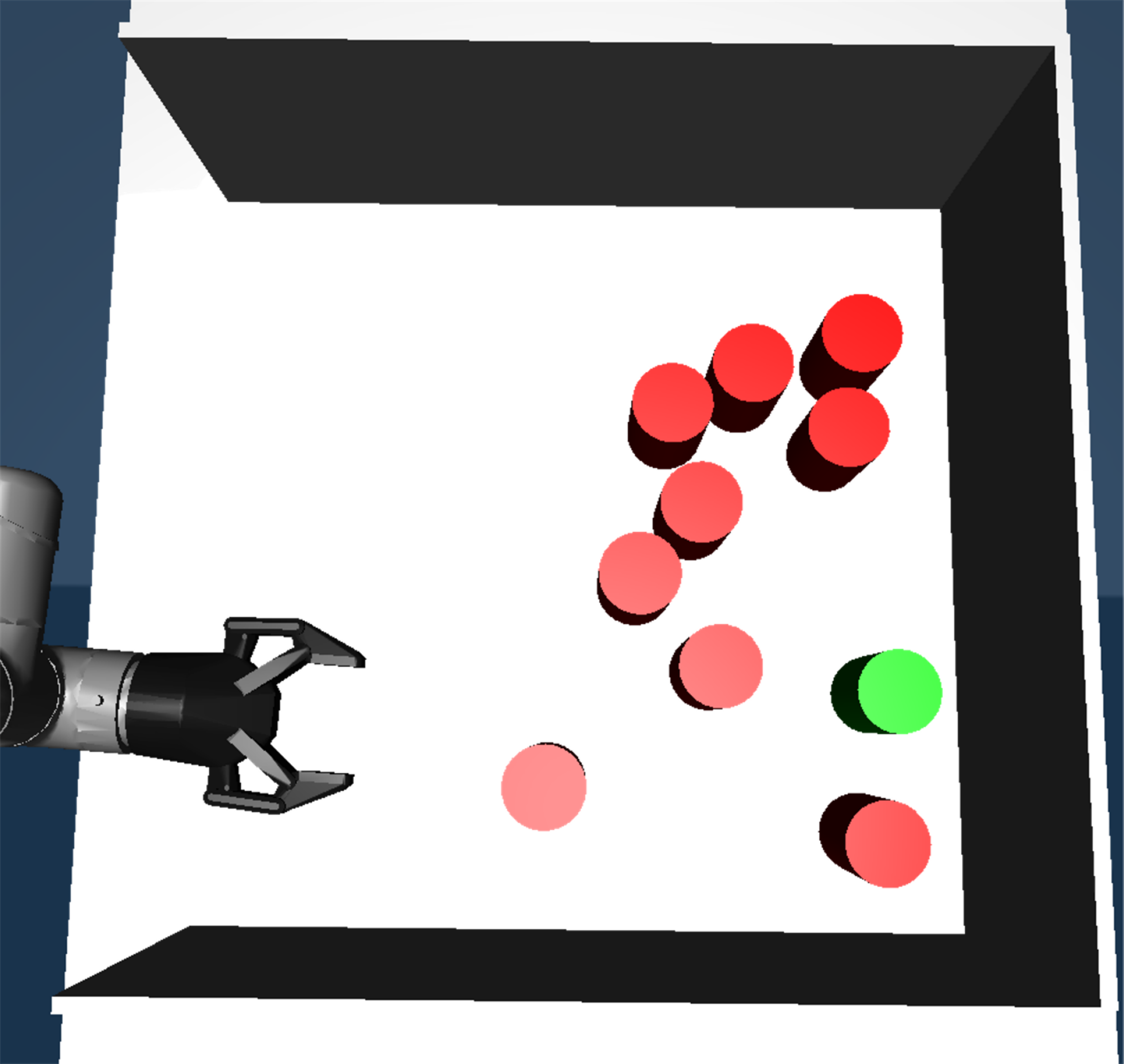
                \caption{}
                \label{fig:heuristic_explained_1}
            \end{subfigure}%
            \begin{subfigure}{.3\linewidth}
                \centering
                \tiny
                \def\svgwidth{0.9\columnwidth}
\begingroup%
  \makeatletter%
  \providecommand\color[2][]{%
    \errmessage{(Inkscape) Color is used for the text in Inkscape, but the package 'color.sty' is not loaded}%
    \renewcommand\color[2][]{}%
  }%
  \providecommand\transparent[1]{%
    \errmessage{(Inkscape) Transparency is used (non-zero) for the text in Inkscape, but the package 'transparent.sty' is not loaded}%
    \renewcommand\transparent[1]{}%
  }%
  \providecommand\rotatebox[2]{#2}%
  \newcommand*\fsize{\dimexpr\f@size pt\relax}%
  \newcommand*\lineheight[1]{\fontsize{\fsize}{#1\fsize}\selectfont}%
  \ifx\svgwidth\undefined%
    \setlength{\unitlength}{950.25bp}%
    \ifx\svgscale\undefined%
      \relax%
    \else%
      \setlength{\unitlength}{\unitlength * \real{\svgscale}}%
    \fi%
  \else%
    \setlength{\unitlength}{\svgwidth}%
  \fi%
  \global\let\svgwidth\undefined%
  \global\let\svgscale\undefined%
  \makeatother%
  \begin{picture}(1,0.94711918)%
    \lineheight{1}%
    \setlength\tabcolsep{0pt}%
    \put(0,0){\includegraphics[width=\unitlength,page=1]{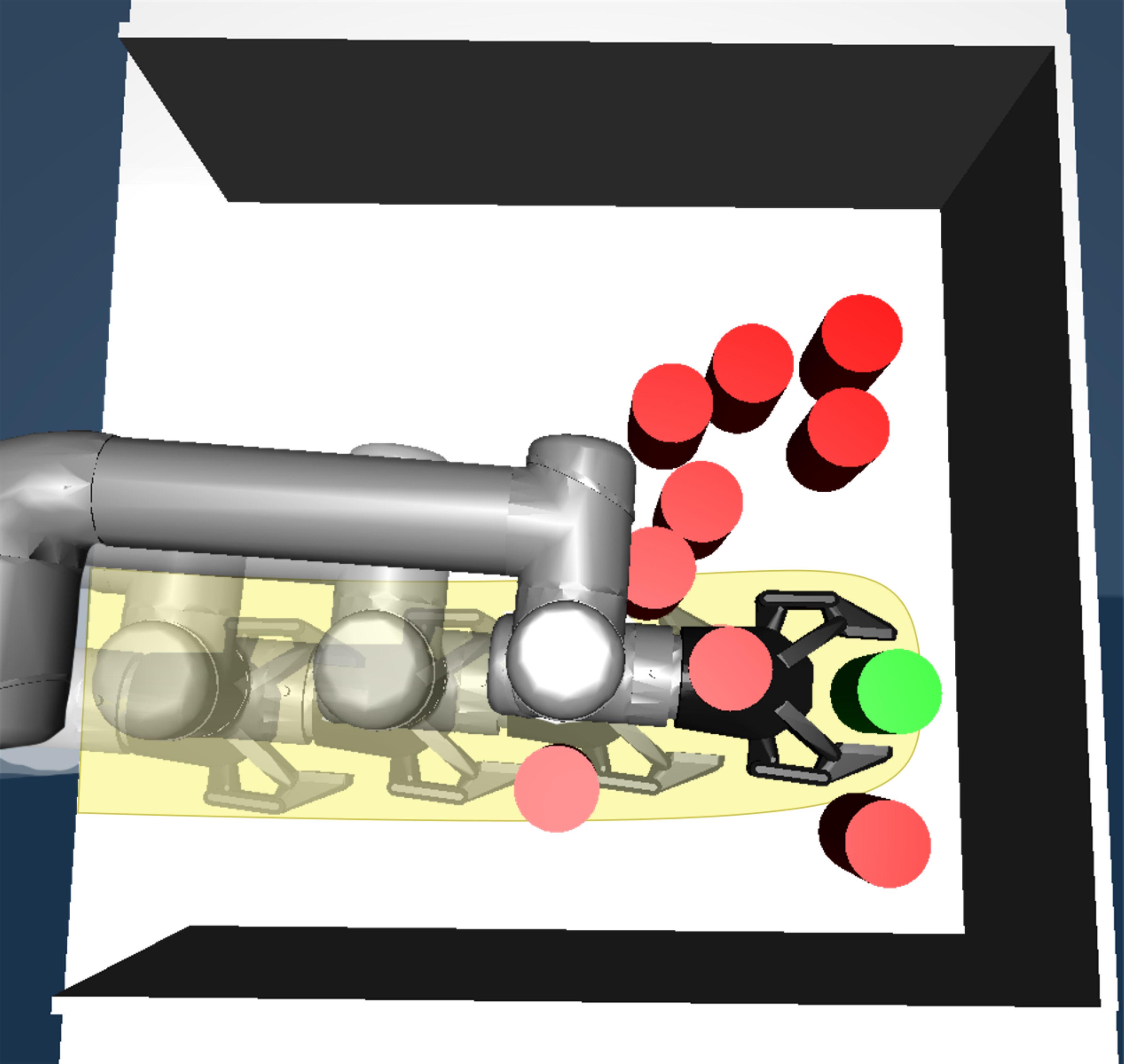}}%
    \put(0.45086655,0.17464552){\color[rgb]{0,0,0}\makebox(0,0)[lt]{\lineheight{1.25}\smash{\begin{tabular}[t]{l}$o_7$\end{tabular}}}}%
  \end{picture}%
\endgroup%

                \caption{}
                \label{fig:heuristic_explained_2}
            \end{subfigure}%
            \begin{subfigure}{.3\linewidth}
                \centering
                \tiny
                \def\svgwidth{0.9\columnwidth}
\begingroup%
  \makeatletter%
  \providecommand\color[2][]{%
    \errmessage{(Inkscape) Color is used for the text in Inkscape, but the package 'color.sty' is not loaded}%
    \renewcommand\color[2][]{}%
  }%
  \providecommand\transparent[1]{%
    \errmessage{(Inkscape) Transparency is used (non-zero) for the text in Inkscape, but the package 'transparent.sty' is not loaded}%
    \renewcommand\transparent[1]{}%
  }%
  \providecommand\rotatebox[2]{#2}%
  \newcommand*\fsize{\dimexpr\f@size pt\relax}%
  \newcommand*\lineheight[1]{\fontsize{\fsize}{#1\fsize}\selectfont}%
  \ifx\svgwidth\undefined%
    \setlength{\unitlength}{950.25bp}%
    \ifx\svgscale\undefined%
      \relax%
    \else%
      \setlength{\unitlength}{\unitlength * \real{\svgscale}}%
    \fi%
  \else%
    \setlength{\unitlength}{\svgwidth}%
  \fi%
  \global\let\svgwidth\undefined%
  \global\let\svgscale\undefined%
  \makeatother%
  \begin{picture}(1,0.94711918)%
    \lineheight{1}%
    \setlength\tabcolsep{0pt}%
    \put(0,0){\includegraphics[width=\unitlength,page=1]{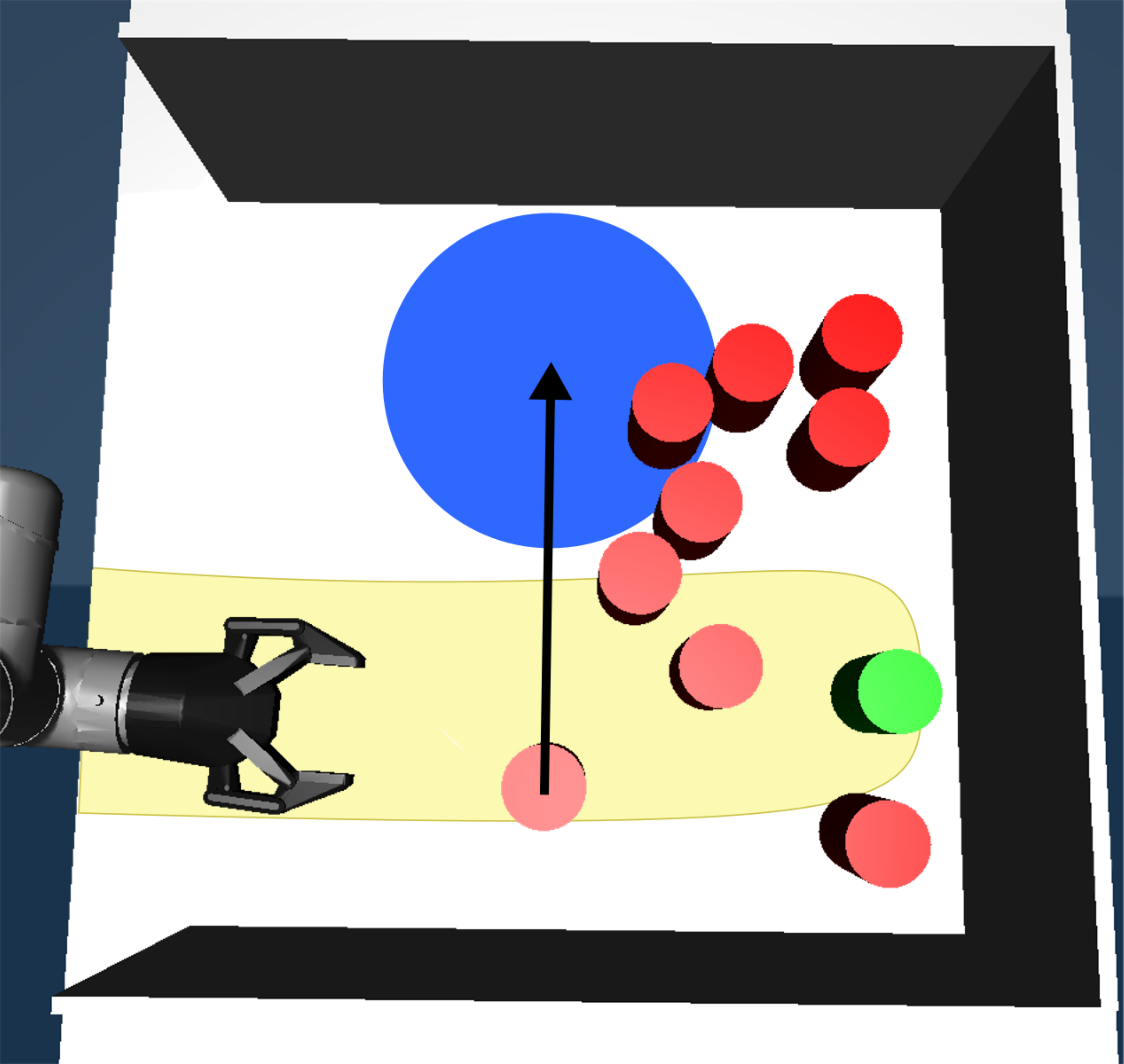}}%
    \put(0.52515689,0.26180397){\color[rgb]{0.48627451,0.44313725,0}\rotatebox{-0.06066708}{\makebox(0,0)[lt]{\lineheight{1.25}\smash{\begin{tabular}[t]{l}$\mathit{V_{swept}}$\end{tabular}}}}}%
  \end{picture}%
\endgroup%

                \caption{}
                \label{fig:heuristic_explained_3}
            \end{subfigure}
            \caption{\heuristicplannername: (a) Initial state.
                (b) The robot moves on a straight line to the goal object, $o_g$, to
                obtain the first blocking obstacle ($o_7$) and the swept volume
                (yellow area). (c) The heuristic produces a high-level action
                for $o_7$ indicated by the arrow and the target region (blue).
                This process is repeated until $V_{swept}$ contains no
                blocking obstacle.}
            \label{fig:heuristic_explained}
        \end{figure}

        \begin{algorithm}[!t]
            \caption{\heuristicplannername Planner}\label{alg:grtc_heuristic}
            \algrenewcommand\algorithmicindent{0.8em}%
            \begin{algorithmic}[1]
                \Function{NextHighLevelAction}{$q_{current}$}
                	\State $o_b \gets$ find the first blocking obstacle to $o_g$ \label{algline:find_first_blocking_obstacle}
					\If{there exists a blocking obstacle $o_b$}
                    	\State $x_b, y_b \gets$ {find collision-free placement of $o_b$} \label{algline:find_region}
                        \State \textbf{return} $o_b, x_b, y_b$ \label{algline:return_action}
					\EndIf
                    \State \textbf{return} $o_g$ \Comment{No blocking obstacle, reach the goal} \label{algline:return_goal}
                \EndFunction
            \end{algorithmic}
        \end{algorithm}

        We present this approach in \cref{alg:grtc_heuristic} and illustrate it in \cref{fig:heuristic_explained}.
        This heuristic assumes the robot moves on a straight line from its current position towards the goal object (\cref{fig:heuristic_explained_2}).
        The first blocking object, $o_b$ on \cref{algline:find_first_blocking_obstacle}, is identified as the next object to be moved.
        During the straight line motion, we capture the robot's swept volume, $V_{swept}$
        (\cref{fig:heuristic_explained_2}). We randomly sample a
        collision-free target region centroid outside $V_{swept}$
        (\cref{alg:grtc_heuristic} \cref{algline:find_region} and
        \cref{fig:heuristic_explained_3}). The object and the centroid are then
        returned as the next high-level action (\cref{alg:grtc_heuristic}
        \cref{algline:return_action}).

        After every high-level action suggested by the heuristic, the Guided-RTC framework
        (\cref{alg:grtc}) plans and executes it and the state of the system is updated
        ($q_{current}$). The heuristic then suggests a new high-level action
        from $q_{current}$ until there is no blocking obstacle (\cref{alg:grtc_heuristic} \cref{algline:return_goal}).

    \section{Experiments \& Results}
    \label{sec:experiment_results}

    For all experiments, we use the Open Motion Planning Library (OMPL)
    \cite{omplSucan2012} implementation of RRT and KPIECE.  We use MuJoCo\footnote{On
    a computer with Intel Core i7-4790 CPU @ 3.60GHz, 16GB RAM.}
    \cite{mujocoTodorov2012} to implement the system dynamics, $f$.
    For all planners, the overall planning time limit, $T_{overall}$, is 300
    seconds, after which it was considered a failure.
    For \ourplanneracronym and \heuristicplannername, $T_{pushing}$ is 10 seconds.

    \subsection{Simulation Results}%
        \label{sub:simulation_results}

        \begin{figure}[!t]
            \vspace*{2mm}
            \captionsetup[subfigure]{aboveskip=1.0pt,belowskip=1.0pt}
            \centering
            \begin{subfigure}{.24\linewidth}
                \centering
                \def\svgwidth{0.85\columnwidth}
                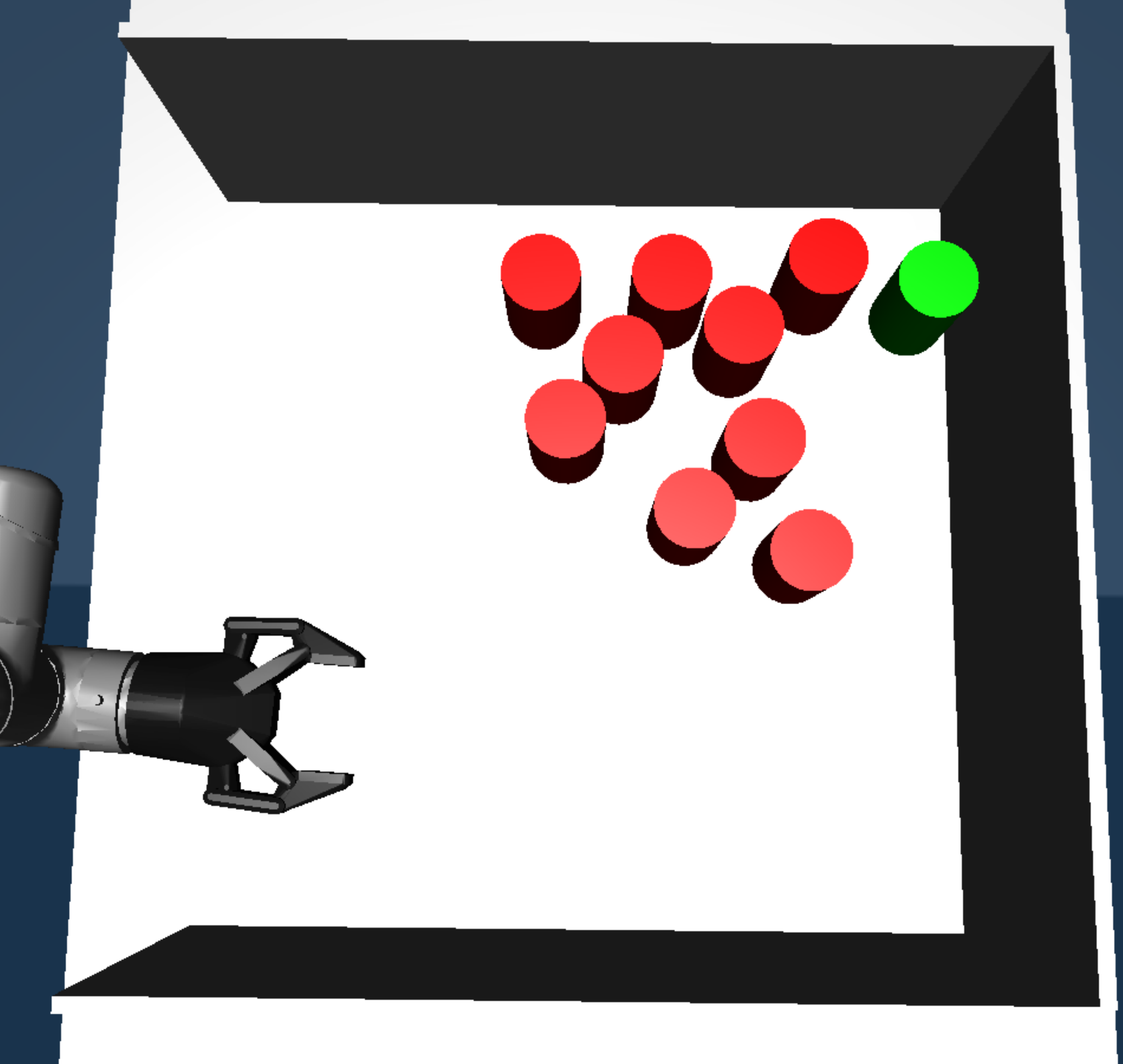
                \caption{S1}
                \label{fig:p1}
            \end{subfigure}%
            \begin{subfigure}{.24\linewidth}
                \centering
                \def\svgwidth{0.85\columnwidth}
                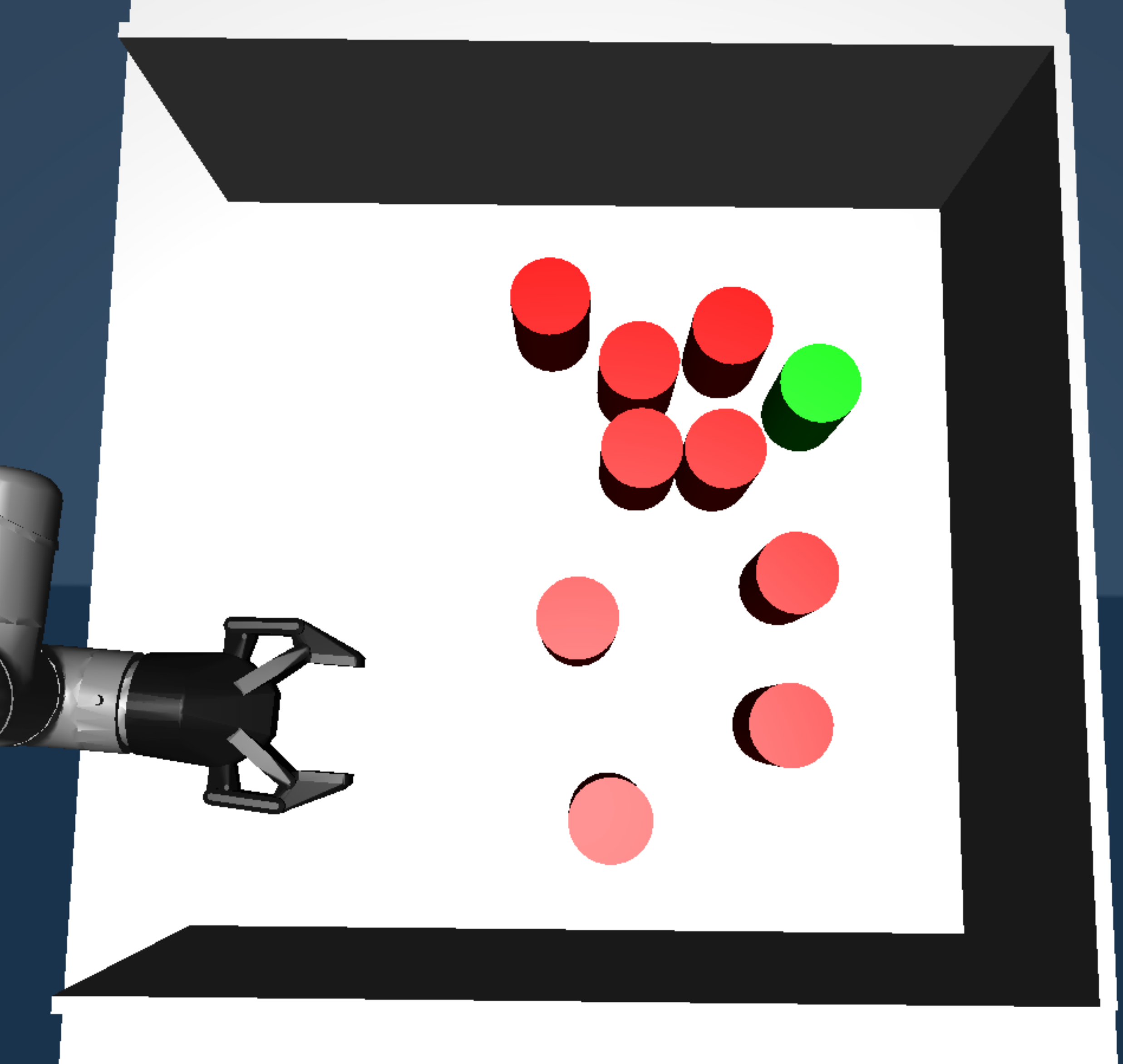
                \caption{S2}
                \label{fig:p2}
            \end{subfigure}%
            \begin{subfigure}{.24\linewidth}
                \centering
                \def\svgwidth{0.85\columnwidth}
                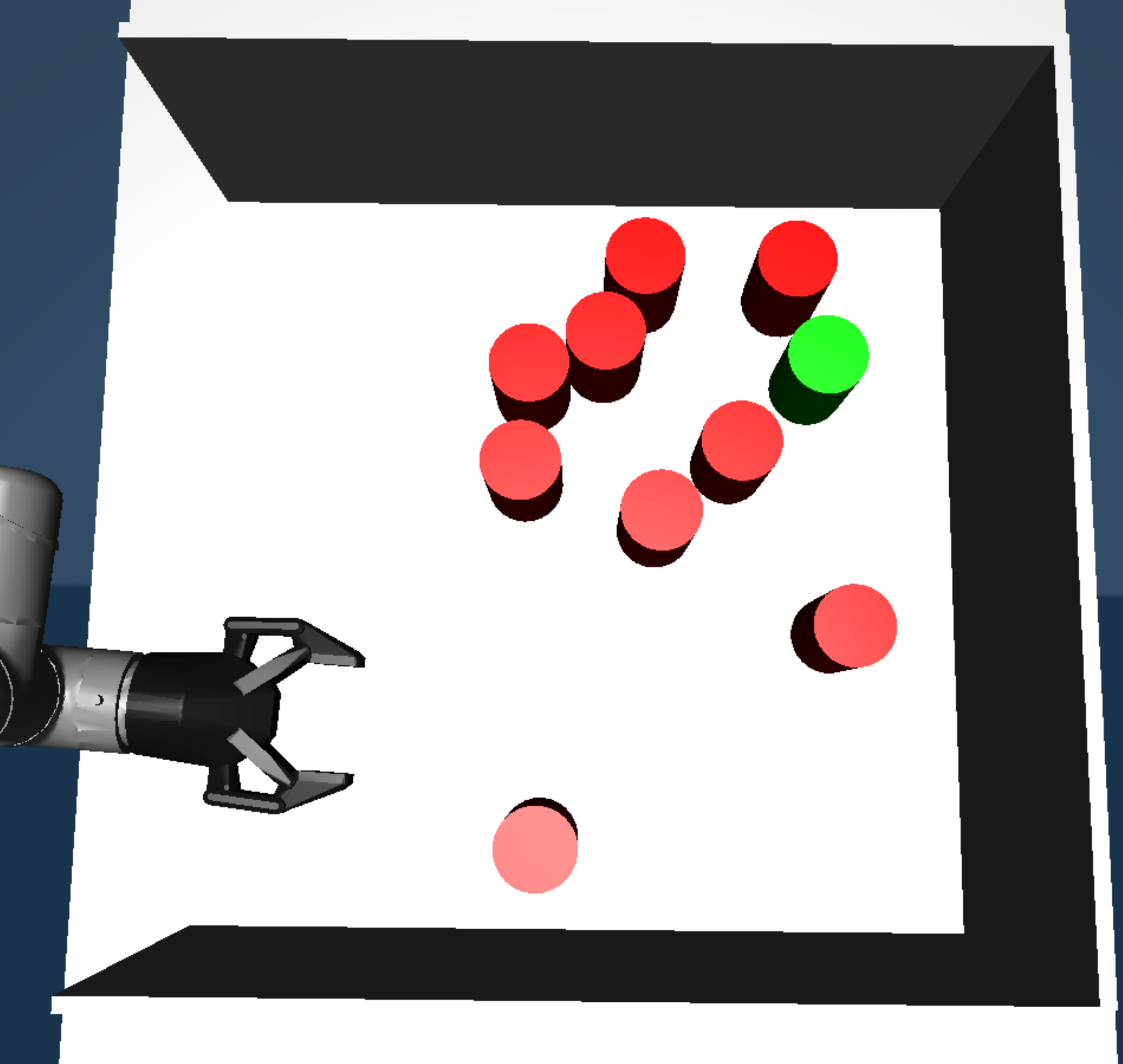
                \caption{S3}
                \label{fig:p3}
            \end{subfigure}%
            \begin{subfigure}{.24\linewidth}
                \centering
                \def\svgwidth{0.85\columnwidth}
                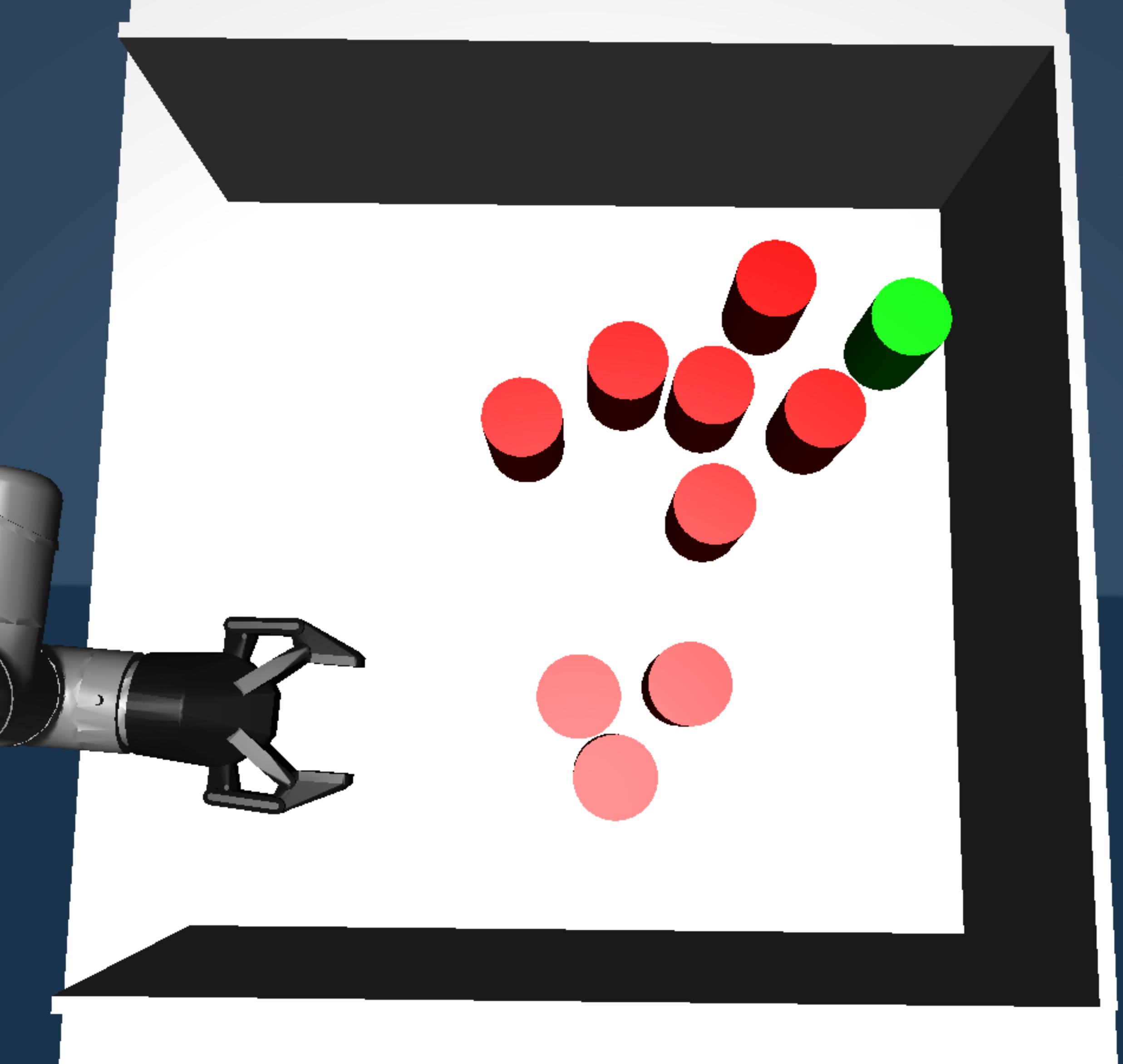
                \caption{S4}
                \label{fig:p4}
            \end{subfigure}
            \caption{Initial states of different problems in simulation
            (S1-S4). Goal object is in green.
            }
            \label{fig:scenes}
        \end{figure}

        We evaluated each approach 100 times by running them 10 times in 10
        different, randomly-generated, scenes. Some of the scenes are presented
        in \crefrange{fig:p1}{fig:p4}. For \ourplanneracronym, the human-operator interacted with each scene
        \textit{once} and from the last state left by the human-operator we ran
        the planner (\cref{alg:grtc} \cref{algline:reach_goal}) to reach for
        the goal object \textit{10 times}.

        \begin{figure}[!t]
            \centering

            \scriptsize
            \def\svgwidth{\columnwidth}
            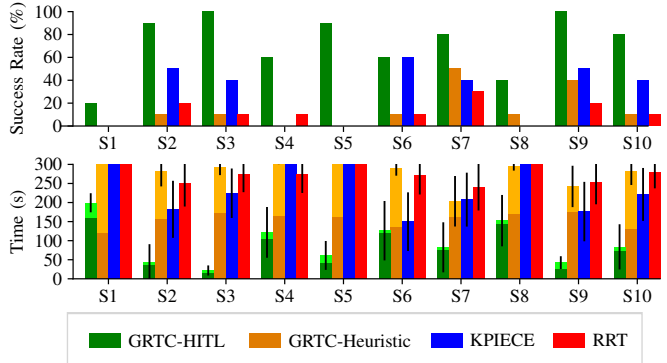

            \caption{Simulation results, for each scene (S1-S10): (Top) Success
            rate. (Bottom) Mean planning time. The error bars indicate
            the 95\% CI. For \ourplanneracronym and \heuristicplannername, the
            dark shade indicates the planning time where the light shade
            indicates the time it took to produce the high-level actions (for
            \ourplanneracronym this is a fraction of the time).}
            \label{fig:simulation_results}
        \end{figure}

        \cref{fig:simulation_results} summarizes the results of our experiments
        for each of the random scenes (S1-S10).
        \cref{fig:simulation_results}-Top shows that \ourplanneracronym yields
        to more successes per scene than any other approach except for S6 which
        was as successful as KPIECE. The overall success
        rate for each approach is 72\% for \ourplanneracronym, 11\% for RRT, 28\% for
        KPIECE and 14\% for \heuristicplannername.
        \cref{fig:simulation_results}-Bottom shows that
        \ourplanneracronym improved the planning time in \textit{all} scenes.

        \cref{tbl:interaction_results} summarizes the guidance performance for
        \ourplanneracronym and \heuristicplannername for all ten scenes.
        Proposed Actions indicates the total number of high-level actions proposed. This number
        includes the successful actions (actions that the planner managed to
        satisfy) and failed actions (actions that the planner could not find
        a solution for). Guidance Time indicates the time spent on generating
        the high-level actions in seconds (in case of
        \ourplanneracronym the time the human-operator was interacting with the
        system and for \heuristicplannername the time took for the heuristic to
        generate the high-level actions).

        \begin{table}[!b]
          \centering
          \caption{Simulation results.}
          \label{tbl:interaction_results}
          \begin{tabular}{|l|cc|cc|}
                    \hline
                    \multirow{2}{*}{} &
                    \multicolumn{2}{c|}{\textbf{\ourplanneracronym}} &
                    \multicolumn{2}{c|}{\textbf{\heuristicplannername}} \\
                    \cline{2-5}
                    & $\mu$ & $\sigma$ & $\mu$ & $\sigma$\\
                    \hline
                    Proposed Actions & 4.9 & 3.3 & 88.4 & 58.2 \\
                    Successful Actions & 3.1 & 1.0 & 3.0 & 1.4 \\
                    Guidance Time (s) & 13.6 & 10.0 & 124.3 & 81.7 \\
                    \hline
                  \end{tabular}
        \end{table}%

        \begin{table}[!b]
          \centering
          \caption{Real-world results.}
          \label{tbl:real_robot_results}
          \begin{tabular}{|l|c|c|c|}
            \hline
            \multirow{1}{1.47cm}{} &
            \multicolumn{1}{c|}{\textbf{\ourplanneracronym}} &
            \multicolumn{1}{c|}{\textbf{KPIECE}} &
            \multicolumn{1}{c|}{\textbf{RRT}}\\
            \hline
            Successes & 7 & 1 & 2\\
            Planning Failures & 2 & 4 & 8\\
            Execution Failures & 1 & 5 & 0\\ \hline
          \end{tabular}
        \end{table}

    \subsection{Real-robot results}%
        \label{sub:real_robot_results}

        We performed experiments using a UR5 manipulator on a Ridgeback
        omnidirectional base. We used the OptiTrack motion capture system to detect
        initial object/robot poses and to update the state in the human interface after
        every high-level action.

        We evaluated RRT, KPIECE and \ourplanneracronym performance in \textit{ten} different
        problems in the real world.

        \cref{tbl:real_robot_results} summarizes the success rate of each
        approach in the real world. When we say that the robot failed during execution,
        we mean that although the planner found a solution,
        when the robot executed the solution in the real-world, it
        either failed to reach the goal
        object, or it violated some constraint
        (hit the shelf or dropped an object to the floor).
        These execution failures were due to the uncertainty in the real world:
        The result of the robot's actions in the real-world yield to different states than the ones
        predicted by the planner.

        The success rate for \ourplanneracronym, RRT and KPIECE is 70\%, 20\%,
        and 10\% respectively. \ourplanneracronym failed 20\% during planning
        and 10\% during execution. KPIECE was more successful during planning
        than RRT but failed most of the times during execution. RRT, on the other,
        hand accounts for more failures during planning than any other approach.

        In \cref{fig:real_experiment_explained_1} we show an example. The human operator
        provides the first high-level action in \cref{fig:real_experiment_explained_1_1}
        and then indicates the goal object in \cref{fig:real_experiment_explained_1_3}
        which is reached in \cref{fig:real_experiment_explained_1_4}.

    \section{Conclusions}
    \label{sec:conclusions}
    We introduced a new human-in-the-loop framework for physics-based
    non-prehensile manipulation in clutter (\ourplanneracronym). We showed through 
    simulation and real-world experiments that \ourplanneracronym is more
    successful and faster in finding solutions than the three baselines we
    compared with.

    \addtolength{\textheight}{-2cm}

    \bibliographystyle{IEEEtran}
    \bibliography{refs}
\end{document}